\documentclass{article}
\usepackage{ijcai23}
\usepackage [table] {xcolor}
\usepackage{times}
\usepackage{soul}
\usepackage{url}
\usepackage[hidelinks]{hyperref}
\usepackage[utf8]{inputenc}
\usepackage[small]{caption}
\usepackage{graphicx}
\usepackage{amsmath}
\usepackage{amsthm}
\usepackage{booktabs}
\usepackage{algorithm}
\usepackage{algorithmic}
\usepackage[switch]{lineno}
\usepackage{enumitem}
\usepackage{color,xcolor}
\usepackage{multirow}
\usepackage{amsthm} 

\usepackage{newfloat}
\usepackage[most]{tcolorbox}
\usepackage{listings}  
\usepackage{multicol}  
\usepackage{amssymb}

\usepackage{amsthm} 

\newtheorem{definition}{Definition}

\DeclareCaptionStyle{ruled}{labelfont=normalfont,labelsep=colon,strut=off} 
\floatstyle{ruled}
\newfloat{listing}{tb}{lst}{}
\floatname{listing}{Listing}

\definecolor{ccr}{RGB}{10,110,150}  

\usepackage{hyperref}
\hypersetup{hypertex=true,
    colorlinks=true,
    linkcolor=ccr,
    anchorcolor=ccr,
    citecolor=ccr}

\title{Exploring Beyond Logits: Hierarchical Dynamic Labeling Based on Embeddings for Semi-Supervised Classification}

\author{
    Anonymous author
    \affiliations
    Submitted to IJCAI 2024
    \emails
    Paper ID: 521
}

\author{
Yanbiao Ma\and
Licheng Jiao\and
Fang Liu\and
LingLing Li\and
Shuyuan Yang\And
Xu Liu
\affiliations
Key Laboratory of Intelligent Perception and Image Understanding of the Ministry of Education\\
International Research Center for Intelligent Perception and Computation, Xi'an 710071, China\\
School of Artificial Intelligence, Xidian University, Xi'an 710071, China\\
\emails
ybmamail@stu.xidian.edu.cn,
lchjiao@mail.xidian.edu.cn
}

\begin{document}

\maketitle

\begin{abstract}
In semi-supervised learning, methods that rely on confidence learning to generate pseudo-labels have been widely proposed. However, increasing research finds that when faced with noisy and biased data, the model's representation network is more reliable than the classification network. Additionally, label generation methods based on model predictions often show poor adaptability across different datasets, necessitating customization of the classification network. Therefore, we propose a Hierarchical Dynamic Labeling (HDL) algorithm that does not depend on model predictions and utilizes image embeddings to generate sample labels. We also introduce an adaptive method for selecting hyperparameters in HDL, enhancing its versatility. Moreover, HDL can be combined with general image encoders (e.g., CLIP) to serve as a fundamental data processing module. We extract embeddings from datasets with class-balanced and long-tailed distributions using pre-trained semi-supervised models. Subsequently, samples are re-labeled using HDL, and the re-labeled samples are used to further train the semi-supervised models. Experiments demonstrate improved model performance, validating the motivation that representation networks are more reliable than classifiers or predictors. Our approach has the potential to change the paradigm of pseudo-label generation in semi-supervised learning.
\end{abstract}

\section{Introduction}

With the advent of the era of large-scale models, researchers have an increasingly urgent demand for data \cite{paper11,paper13,paper39}. Manual annotation stands out as a reliable method to enhance label quality, but its cost becomes prohibitively expensive for large-scale datasets. Current automatic labeling methods mainly focus on utilizing Deep Neural Networks trained on labeled data to predict labels for unknown samples. Particularly in the field of semi-supervised learning, Confident Learning \cite{paper1} has become a ubiquitous strategy for obtaining pseudo-labels for samples \cite{paper14,paper9,paper10}. Specifically, the approach involves training a model with labeled data while utilizing the model to generate pseudo-labels for unlabeled samples with high confidence. As the model's performance improves, more samples are selected based on high confidence, and semi-supervised learning benefits from this iterative optimization process. However, methods based on Confidence learning face two primary limitations: 

\begin{tcolorbox}[
  enhanced, 
  frame hidden, 
  borderline west={1.5mm}{-0.5pt}{pink}, 
  colback=pink!15, 
  sharp corners, 
  boxrule=0pt, 
  left=-1.5mm, 
  right=0mm, 
  top=0mm, 
  bottom=0mm, 
]
\begin{itemize}[]
\item Generating prediction confidence requires training classification networks specifically for different datasets, making them less versatile.
\item When labeled data is biased or contains noise (i.e., samples with incorrect labels), classifiers in deep neural networks tend to exhibit significant bias \cite{paper18,paper31} and excessive memorization of noisy samples \cite{paper15,paper16}, leading to the mislabeling of unknown samples.
\end{itemize}
\end{tcolorbox}

\textbf{\emph{Motivation}} A promising annotation approach is one that does not rely on the model's predictions. Recent studies have found that in real-world scenarios, the representation network of a model appears to be more reliable than prediction networks or classifiers \cite{paper19,paper21}. In the realm of long-tailed image recognition, the motivation behind decoupled training \cite{paper20} is rooted in the observation that biases in models stem from classifiers, whereas representation networks can obtain unbiased image embeddings. In the domain of noise detection, it has been observed that models tend to excessively memorize noisy samples, hindering accurate noise identification based on the model's prediction confidence \cite{paper15}. The aforementioned studies suggest that, compared to methods relying on model predictions, developing embedding-based, data-centric approaches can avoid many shortcomings introduced by the learning process \cite{paper33}. Therefore, our focus is on developing an embedding-based image annotation method to enhance tasks like semi-supervised learning, leveraging unlabeled data to improve model performance. The main contributions of this paper are summarized as follows:

\begin{itemize}[]
\item We propose a Hierarchical Dynamic Labeling algorithm (HDL) that does not rely on model predictions (see Section \ref{sec4.3}). This method is highly flexible and can also be combined with CLIP to serve as a general-purpose data preprocessing module (see Section \ref{sec5.7}).
\item We propose a method for adaptively selecting the hyperparameter $k$ for HDL, making HDL less reliant on empirical choices and thereby enhancing its universality.
\item In both class-balanced (see Section \ref{sec5.4}) and class-imbalanced scenarios (see Section \ref{sec5.5}), our method significantly enhances the performance of semi-supervised models (Table \ref{table2} and \ref{table3}), validating the notion that the representation network is more reliable than classifiers or predictors. Furthermore, our approach has the potential to shift the paradigm of generating pseudo-labels in semi-supervised learning. (see Section \ref{sec5}).
\end{itemize}

\section{Related Works}
\textbf{\emph{Generating image pseudo-labels based on confidence learning}}
The core idea of confidence learning \cite{paper1} is to identify labeling errors in a dataset by the predictive confidence of the model. Specifically, when the prediction confidence of a sample is lower than a predetermined threshold, the sample is considered to be mislabeled with a high probability. In order to make the recognition results more reliable, it becomes a popular approach to use multiple models for joint prediction. The idea of confidence learning is widely used in the field of semi-supervised learning for the generation of image pseudo-labels \cite{paper2,paper3,paper4,paper5}. The basic paradigm is to train on labeled data first and then predict the pseudo-labels of unlabeled data, which in turn selects high-confidence pseudo-labels to train the model. For example, MixMatch \cite{paper3} augments unlabeled samples $k$ times and then uses the model to make $k$ predictions on the augmented samples and averages them to obtain pseudo-labels. In order to improve the reliability of model predictions, ReMixMatch \cite{paper4} adds a distribution alignment step to MixMatch with the aim of making the model predicted labels closer to the existing label distribution.FixMatch \cite{paper6} uses weak data augmentation to obtain pseudo-labels for the samples and is used to supervise the output of strong data augmentation. The recent Dash \cite{paper7}, FlexMatch \cite{paper8} and FreeMatch \cite{paper9} enhance the quality of pseudo-labels by designing a better method of choosing the prediction confidence threshold. 
\textbf{Different from them}, we label unknown samples based on the representation of images and justify our motivation by improving existing methods.

\section{Preliminary}
\label{sec3}

\subsection{Task and Symbol Definition}
\label{sec3.1}

Given a labeled dataset $D: =\{(x_n,y_n)\}_{n\in [N]}$, where $[N]:=\{1,2,\dots,N \}$ and $y_n\in \{1,2,\dots,C \}$. Additionally, we have an unlabeled dataset $D': =\{x'_m\}_{m\in [M]}$, where $y'_m$ represents the potential label corresponding to $x'_m$. There are two possible scenarios for the value of $y'_m$: (1) $\forall m\in[M],y'_m\in [C]$, and (2) $\exists m\in[M],y'_m\notin [C]$. We focus on closed-set image labeling, so the label spaces of datasets $D$ and $D'$ are consistent, i.e., $\forall m\in[M],y'_m\in\{1,2,\dots,C\}$.

\subsection{Label Clusterability}
\label{sec3.2}

This work aims to achieve automatic image annotation without training, and the label clusterability \cite{paper15,paper35} provides a basis for our research. Intuitively, clusterability means that two embeddings that are closer have a higher probability of belonging to the same true class. Label clusterability can be formally defined as follows.

\begin{definition}[$(k,\delta_k)$ Label Clusterability]
\label{def3}
Given a dataset $D$ and an image encoder $f(\cdot)$, extract the embeddings $X=f(D)$ of all images in $D$. For $\forall x\in X$, if the probability that $x$ and its $k$ nearest neighbors belong to the same true class is at least $1-\delta_k$, then the dataset $D$ is said to satisfy $(k,\delta_k)$ label clusterability. When $\delta_k=0$, the dataset $D$ is said to satisfy $k$-NN label clusterability.
\end{definition}

$\delta_k$ represents the probability that the dataset $D$ violates the clusterability, which is related to the size of $k$ and the quality of the embedding. It is obvious that $\delta_k$ increases as $k$ increases. This is because when $k$ is larger, the $k$ nearest neighbors of the central instance are more likely to contain embeddings of other categories. When the quality of the embedding is lower, it means that the image encoder has a worse ability to distinguish between categories, which may lead to confusion between embeddings of different categories, making $\delta_k$ larger. In the context of automatic image annotation in general scenarios, large-scale visual pre-training models with powerful visual representation capabilities, such as CLIP \cite{paper38}, have emerged as a compelling choice for extracting embeddings.

\section{Automatic classification using embedding}
\label{sec4}

In this section, we first introduce the simplest idea of utilizing label clusterability for annotation, and then progressively refine it to propose a more general and superior algorithm. For the sake of clarity in presentation, CLIP is utilized in this section to extract embeddings from the dataset; however, the experimental section will incorporate a variety of image encoders for a more extensive evaluation.

\subsection{Intuition for image labeling relying on $(k,\delta _k)$ label clustrability}
\label{sec4.1}

Given a set of labeled embeddings $D: = \{(x_n,y_n) \}_{n\in [N]}$ and a set of unlabeled embeddings $D': = \{x'_m \}_{m\in [M]}$, all embeddings are extracted by the CLIP. Assuming that $D$ satisfies the $k_1$-NN label clustrability, this means that we can determine the label of the central instance by using the labels of its $k_1$ nearest neighbors. Specifically, for the embedding $x'_m,m\in [M]$ to be labeled, we select $k_1$ nearest neighbors $x'_{m_1}, x'_{m_2}, \dots, x'_{m_{k_1}}$ from $D$, and their corresponding labels are $y'_{m_1}, y'_{m_2}, ..., y'_{m_{k_1}}$. We convert all labels into the form of one-hot encoding, and then the label of $x'_m$ can be obtained by voting on its $k_1$ nearest neighbors, which is
\begin{equation}
\begin{split}
y'_m=argmax_{i\in [C]}(\frac{1}{k_1}\sum_{j}^{k_1}y'_{m_j}  )[i],y'_{m_j}\in \mathbb{R}^C.
\end{split}
\end{equation}

The above process seems to be a valuable method for image annotation using clusterability, but this method has high requirements for clusterability and there is room for optimization. Let's analyze it below.
Although the embeddings in $D'$ have no label information, their label space is consistent with $D$. Therefore, after mixing the set $D'$ of embeddings extracted by CLIP with $D$, all the embeddings still satisfy a certain degree of clustering. When selecting $k_1$ nearest neighbors for $x'_m$, the unlabeled embeddings are excluded, but these unlabeled embeddings may be closer to the central instance $x'_m$. Let the distance between the $k_1$th nearest neighbor of $x'_m$ and $x'_m$ be $d_{k_1}$, and simultaneously select $k_1$ nearest neighbors for $x'_m$ from $D'$ and $D$. Let the distance between the $k_1$th nearest neighbor and $x'_m$ be $d_{k_2}$. Clearly, $d_{k_1}\ge d_{k_2}$. The above analysis shows that methods that do not use unlabeled embeddings for voting have higher requirements for label clusterability. In order to make the automatic labeling method based on embeddings more general and universal, we pursue the use of unlabeled embeddings to assist in labeling, thereby reducing the requirements of the labeling algorithm for label clusterability and improving the performance of the labeling algorithm.

\begin{algorithm*}[t]
\caption{Hierarchical Dynamic Labeling (HDL)}
\label{alg1}
\textbf{Input}: $k$, labeled embedding set $D: = \{(x_n,y_n) \}_{n\in [N]}$, unlabeled embedding set $D': = \{x'_m \}_{m\in [M]}$.\\
\textbf{Output}: $D$
\begin{algorithmic}[1] 
\REPEAT{  }
\STATE For each embedding in $D'$, we search its $k$ nearest neighbors from both $D$ and $D'$, count the number of embeddings in the $k$ nearest neighbors that belong to $D$, and represent it as $L_1,L_2,\dots,L_M$.
\STATE $First\_level\_data = []$     \quad \quad  \textcolor{blue}{\# $First\_level\_data$ is used to store the embeddings of the first level.}
\FOR{$m = 1$ to $M$}
\IF{$L_m==np.max(L_1,L_2,\dots,L_M)$}
\STATE $First\_level\_data.append(x'_m)$   \quad \quad  \textcolor{blue}{\# Save the eligible embedding into $First\_level\_data$.}
\ENDIF
\ENDFOR
\STATE $FLD_{number} = len(First\_level\_data),  \quad order = torch.zeros(FLD_{number})$
\FOR{$i = 1$ to $FLD_{number} $}
\STATE \textcolor{blue}{\# $Find\_kNN(\cdot)$ for finding $k$ nearest neighbors.}
\STATE $First\_level\_data[i-1]_1,\dots,First\_level\_data[i-1]_k = Find\_kNN(First\_level\_data[i-1])$
\STATE \textcolor{blue}{\# Voting using labeled $L_m==np.max(L_1,L_2,\dots,L_M)$ nearest neighbors.}
\STATE $y(First\_level\_data[i-1]) = Vote(First\_level\_data[i-1]_1,\dots,First\_level\_data[i-1]_k)$
\STATE $D.append(First\_level\_data[i-1], y(First\_level\_data[i-1]))$
\STATE Find the $k$ nearest neighbors for each embedding in $First\_level\_data$ except for $First\_level\_data[i-1]$, and count the number of nearest neighbors that belong to $D$ for each embedding, denoted as $l_m,m\in[FLD_{number}],m\ne i-1$.
\STATE Remove $(First\_level\_data[i-1], y(First\_level\_data[i-1]))$ from $D$.
\STATE $order[i-1] = sum(l_m,m\in[FLD_{number}],m\ne i-1)$
\ENDFOR
\STATE $\_,index = sort(order)$   \quad \quad  \textcolor{blue}{\# The labeling order of the $FLD_{number}$ embeddings is stored in index.}
\FOR{$i = 1$ to $FLD_{number}$}
\STATE Label $First\_level\_data[index[i-1]]$ and move into $D$ after labeling.
\STATE Remove $First\_level\_data[index[i-1]]$ from $D'$.
\ENDFOR
\UNTIL{$|D'|=0 $} \quad \quad  \textcolor{blue}{\# $|D'|$ denotes the number of embeddings in $D'$}
\end{algorithmic}
\end{algorithm*}

\begin{listing*}[t]%
\caption{Adaptive selection of k}%
\label{alg2}
\begin{lstlisting}[language=Python] 
def auto_k(clip_feature_labeling, Label, p=0.1, k_upper_limit = 20):  
	PBTY = torch.zeros(k_upper_limit)    #PBTY is used to store the probability that clustrability holds.
	I = torch.zeros(k_upper_limit)  
    for i in tqdm(range(1,k_upper_limit)):
        features = clip_feature_labeling  
        m = int(features.shape[0]*p)  
        selected_features=features[np.random.randint(0,features.shape[0],(m,))]  
        norms = torch.norm(selected_features, dim=1)
        selected_features = selected_features / norms.view(-1,1)
        dist_matrix = 1 - torch.matmul(selected_features, features.T) 
        _, indices = dist_matrix.topk(i+1, dim=1, largest=False)  
        Y = Label[indices]  
        row_indices = list(range(Y.shape[0]))  
        result = list(map(lambda idx: check_row_equal(Y, idx), row_indices)) 
        probability = sum(result) / Y.shape[0]  
        PBTY[i-1] = probability
        k1 = math.ceil((i+1)/2)-1 
        I[i-1] = scipy.special.betainc(i+1-k1, k1+1, 0.85) 
    max_value, max_index = torch.max(PBTY*I,0)
    return max_index+1
\end{lstlisting} 
\end{listing*}

\subsection{Unlabeled embeddings can promote labeling}
\label{sec4.2}

In the following, we refer to the method introduced in Section \ref{sec4.1} as kNN-DV, and assume that all embedding sets satisfy the $k$-NN label clusterability. Introducing the unlabeled embedding set $D'$ into the selection range of $k$-NN means that there may be more similar embeddings inside the hypersphere $B(x'_m,R)$ centered at $x'_m$ and with radius $R$, compared to kNN-DV. In this case, if we keep the number of nearest neighbors consistent with the base algorithm, we can find enough similar embeddings within a hypersphere centered at $x'_m$ with a volume smaller than $B(x'_m,R)$. 
The aforementioned concept allows for the reduction in the requirement of label clusterability for the embedding set while maintaining a constant number of nearest neighbors.

\begin{figure}[h]
\vskip -0.05in
\centering
\includegraphics[width=3.3in]{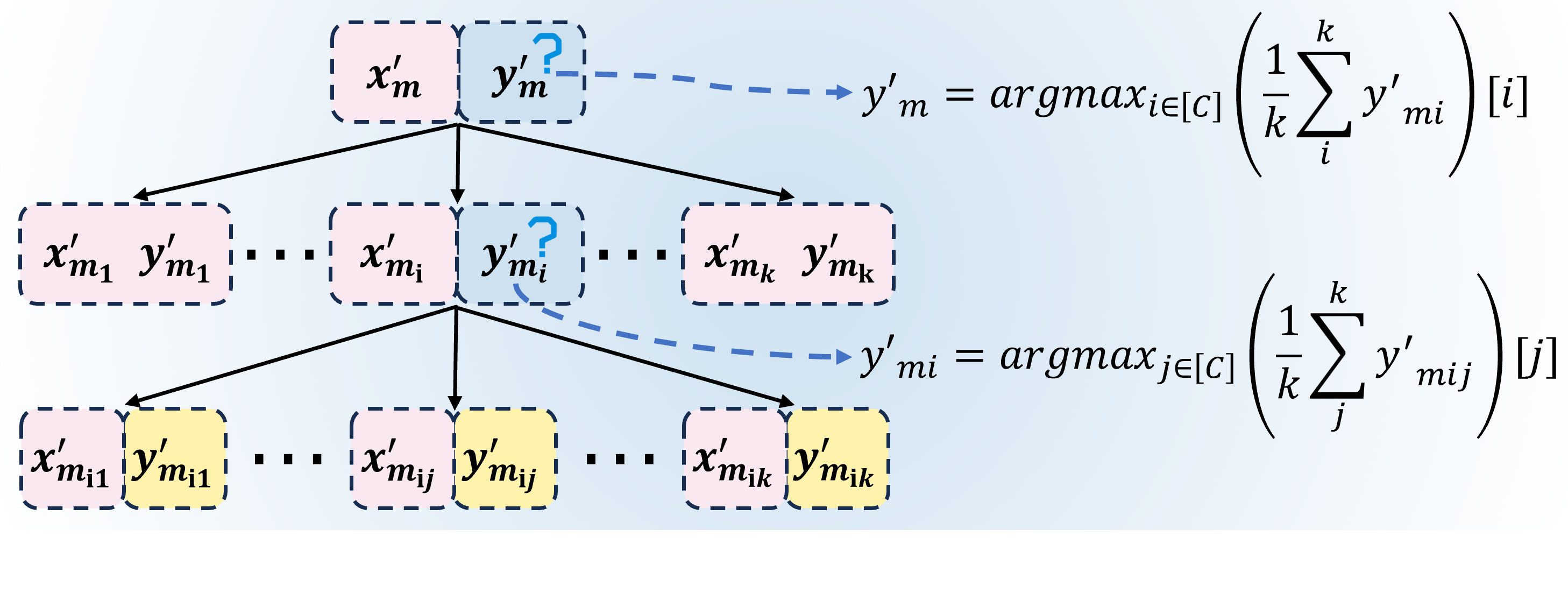}
\vskip -0.1in
\caption{An example of a two-tier tree structure used for label searching. Blue boxes indicate the absence of a label, while yellow boxes represent uncertainty regarding the existence of a label.}
\label{fig6}
\end{figure}

However, we \textbf{face a problem} that must be solved: some of the $k$ nearest neighbors of $x'_{m_1},x'_{m_2},\dots,x'_{m_k}$, may be unlabeled embeddings, i.e., there exists $j\in [k]$ such that $x'_{m_j}\in D'$, which makes it difficult to directly determine label of $x'_m$ through voting. However, we found that the labels of each embedding can be gradually determined through depth-first search of the tree structure. As shown in Figure \ref{fig6}, assume that among $k$ nearest neighbors of $x'_m$, $x'_{m_j}\in D',j\in [k]$, then it is necessary to further find $x'_{m_j}$'s $k$ nearest neighbors. If all of $x'_{m_j}$'s $k$-NN carry label information, or a certain proportion of nearest neighbors carry label information, then use these labels to vote for the category of $x'_{m_j}$, and then determine the category of $x'_m$. If most of $x'_{m_j}$'s $k$ nearest neighbors do not have labels, continue depth-first search. Each step of the deep search satisfies the criterion of label clusterability with reduced requirements.

The intuitive understanding of using tree-based depth-first search for labeling is as follows: given the set $D'$ of embeddings to be labeled, labeled the center instance with the largest number of labeled embeddings in its $k$-NN, then add the labeled embeddings to set $D$ and use them to label other embeddings. That is, the elements of $D$ are dynamically increased during the labeling process until all embeddings belonging to $D'$ have been transferred to $D$. However, the disadvantage of depth-first search based on tree structure is also obvious, such as may fall into circular search, multiple branches repeat search and so on. Therefore, we propose a hierarchical dynamic labeling algorithm.

\subsection{Hierarchical Dynamic Labeling (HDL)}
\label{sec4.3}

Based on the analysis in section \ref{sec4.2}, introducing unlabeled embeddings into the automatic labeling process can reduce the algorithm's requirements for label clusterability. However, using tree-based search and voting has significant drawbacks, so we propose hierarchical dynamic labeling to achieve the same goal as tree-based automatic labeling.

Given a labeled embedding set $D: = \{(x_n,y_n) \}_{n\in [N]}$ and an unlabeled embedding set $D': = \{x'_m \}_{m\in [M]}$, \textbf{the core idea} of hierarchical dynamic labeling with respect to $D'$ involves two points: \textbf{(1)} prioritize labeling the center instance with the largest number of labeled embeddings among its $k$ nearest neighbors; and \textbf{(2)} dynamically update $D$ by transferring newly labeled embeddings to $D$ in real-time.

Why is our approach called hierarchical? This comes from our first core idea. Specifically, for each embedding in $D'$, we search its $k$ nearest neighbors from both $D$ and $D'$, count the number of embeddings in the $k$ nearest neighbors that belong to $D$, and represent it as $L_1,L_2,\dots,L_M$. First, we select all embeddings that satisfy $L_m=\max \{L_1,L_2,\dots,L_M\},m\in [M]$ for labeling, and represent the selected embeddings as $x'_{max,1},\dots,x'_{max,s}, s\le M$. Furthermore, we need to determine the labeling order of the selected $s$ embeddings. Obviously, during the entire labeling process, the labeling order of the embeddings is gradually determined by two levels. In the first level, $s$ samples with priority labeling are determined, and in the second level, the labeling order of the $s$ samples is further determined. Next, we will introduce how to determine the labeling order of the $s$ samples in the second level.

Given $s$ unlabeled embeddings $x'_{max,1},\dots,x'_{max,s}$. Assume that the labeling of $x'_{max,i},i\in [s]$ is first conducted, then $x'_{max,i}$ is transferred to $D$ after the labeling is completed. Then we count the number $[l_{i,1},\dots,l_{i,i-1},l_{i,i+1},\dots,l_{i,s}]\in \mathbb{R}^{s-1}$ in the $k$ nearest neighbors of the remaining $s-1$ embeddings that belong to $D$. Perform the above operation for all $s$ embeddings and place the statistical results for the $i$-th embedding in the $i$-th row of matrix $l$. $l$ is represented as
\begin{equation}
\begin{split}
l=\begin{bmatrix}
  l_{1,2}&  \cdots &l_{1,i}  &l_{1,i+1} &\cdots &l_{1,s} \\
  \vdots & \ddots  &\vdots  &\vdots  &\ddots  & \vdots  \\
  l_{i,1}& \cdots  &l_{i,i-1}  &l_{i,i+1}  & \cdots & l_{i,s} \\
  \vdots & \ddots  &\vdots  &\vdots   &\ddots    &\vdots \\
  l_{s,1}& \cdots  &l_{s,i-1} &l_{s,i} &\cdots    & l_{s,s-1}
\end{bmatrix}\in \mathbb{R}^{s\times (s-1)}.
\nonumber
\end{split}
\end{equation}
Let $l[i]$ represent the $i$-th row of matrix $l$. We sum each row of the matrix to obtain $sum(L) = [sum(l[1])= {\textstyle \sum_{j\ne 1}^{s}l_{1,j}},\dots,sum(l[s])= {\textstyle \sum_{j\ne s}^{s}l_{s,j}} ] \in \mathbb{R}^s$. $sum(l[i])$ reflects the number of embeddings in the $k$ nearest neighbors of the remaining $s-1$ embeddings that belong to $D$ after the $i$-th embedding is labeled. Obviously, a larger $sum(l[i])$ is preferred, so we sort the elements of $sun(L)$ in descending order and prioritize labeling embeddings with larger element values. At this point, we have introduced the proposed hierarchical dynamic labeling algorithm in its entirety. The detailed calculation steps are shown in Algorithm \ref{alg1}.

The core parameter of the hierarchical dynamic Labeling (HDL) algorithm is the number of nearest neighbors $k$. How to adaptively choose $k$ is the key to enhancing the generality of our method. In the next subsection, we will give an adaptive selection method for $k$ in terms of the probability that clustrability holds and the probability that the $k$-nearest-neighbor vote is successful.

\subsection{Adaptive selection and analysis of $k$}
\label{sec4.4}

When $k$ is larger, the majority voting method is obviously more robust and reliable. However, it has to be considered that as $k$ increases, the probability of satisfying the clusterability of the embedding decreases. Therefore, the selection of $k$ value needs to balance the probability of satisfying the clusterability and the advantage brought by majority voting. Given a set of embeddings, assuming the probability of satisfying the clusterability is $\mu_k$, then the probability of correctly voting \cite{paper15} by $k$ nearest neighbors satisfies
\begin{equation}
\begin{split}
P(k)\ge \mu_k\cdot I_{1-e}(k+1-k',k'+1),
\end{split}
\end{equation}
where $k'=\left \lceil (k+1)/2 \right \rceil -1$ and $e$ represents the probability of incorrect labels in the embedding set.
$I_{1-e}(k+1-k',k'+1)$ is a regularized incomplete beta function, calculated as $I_{1-e}(k+1-k',k'+1)=\frac{(k+1)!}{(k-k')!k'!}\int_{0}^{1-e}t^{k-k'}(1-t)^{k'}dt$. As $k$ increases, $\mu_k$ usually decreases while $I_{1-e}(k+1-k',k'+1)$ increases. $P(k)$ can be used as a tool for selecting appropriate $k$ values, where $\mu_k$ is affected by the quality of the embedding, and thus the $\mu_k$ corresponding to different $k$ values need to be statistically obtained from the embedding set. Specifically, a certain proportion of center instances are randomly selected from the labeled embedding set, and their $k$ nearest neighbors are found. The condition of whether the center instances and their $k$ nearest neighbors have the same label is checked, and the proportion of center instances that satisfy this condition is counted to estimate $\mu_k$. The steps for estimating $\mu_k$ and selecting $k$ values are given in Listing \ref{alg2}.

\section{Experiments}
\label{sec5}

In this section, we introduce the experiments from the following three aspects: (1) Experimental objectives and the design of the experimental program. (2) The dataset used in the experiments and the related parameter settings. (3) Experimental results and analysis.

\subsection{Experimental objectives and program design}
\label{sec5.1}

Looking back at the fundamental motivation of this work again: \textbf{image embeddings are more reliable than model predictions}. Therefore, we propose the HDL algorithm based on image embeddings. The goal of this paper is to prove that HDL is superior to labeling methods based on confident learning. Semi-supervised learning with pseudo-labels provides us with a good experimental setting for this purpose. Specifically, the semi-supervised model trains the model while predicting pseudo-labels for unknown samples to achieve alternating optimization. \textbf{This leads to the question:} if HDL based on embeddings is more reliable than model predictions, then after semi-supervised learning convergence, using HDL to relabel unknown samples and continue training the semi-supervised model will further improve performance. If performance cannot be improved, it proves that our method does not bring any extra benefits. Based on this thinking, we extracted image embeddings from a converged semi-supervised model and conducted the above experiments in scenarios with class balance (see Section \ref{sec5.4}) and long-tailed distribution (see Section \ref{sec5.5}).

\subsection{Datasets and Implementation Details}
\label{sec5.2}

We evaluated the performance of HDL in improving semi-supervised methods on both class-balanced and class-imbalanced datasets. The balanced datasets included CIFAR-10, CIFAR-100, and STL-10, while the class-imbalanced dataset was CIFAR-10-LT with various imbalance factors.

\textbf{Class-balanced datasets: CIFAR-10, CIFAR-100, STL-10.} CIFAR-10 \cite{paper22} consists of $60,000$ images across $10$ classes, of which $50,000$ images are used for training and $10,000$ images are used for testing. We follow the common practice in semi-supervised learning to partition the data as follows: for each class, we randomly select $4$, $25$, and $400$ samples as labeled data, and treat the remaining samples in the training set as unlabeled data. CIFAR-100 \cite{paper22} comprises $100$ classes with $500$ training samples and $100$ testing samples per class. We randomly select $25$ and $100$ samples as labeled data for each class, and consider the remaining samples as unlabeled data. STL-10 \cite{paper23} includes $10$ classes with $500$ training samples, $800$ testing samples, and an additional $10,000$ unlabeled samples. For each class, we randomly select $100$ samples as labeled data from the training set.

\begin{table*}[t]
\caption{Error rates (mean $\pm$ std $\%$) on CIFAR-10, CIFAR-100, and stl-10 datasets. The best results are highlighted in \textbf{bold}, and the second-best results are \underline{underlined}. Note that we evaluated our approach using $5$ experiments with different random seeds.}
\label{table2}
\vskip -0.05in
\centering  
\renewcommand\arraystretch{1.07}
\setlength{\tabcolsep}{10.1pt} 
\begin{tabular}{l|ccc|cc|c}
\hline \toprule
\multirow{2}{*}{Method}  & \multicolumn{3}{c}{CIFAR-10}         & \multicolumn{2}{c}{CIFAR-100} & STL-10      \\ \cline{2-7} 
                         & 40 labels & 250 labels & 4000 labels & 2500 labels   & 10000 labels  & 1000 labels \\ \hline
$\Pi$-Model & 74.34$\pm$1.76     & 54.26$\pm$3.97      & 41.01$\pm$0.38       & 57.25$\pm$0.48         & 37.88$\pm$0.11         & 32.78$\pm$0.40       \\
Pesudo-Labeling          &74.61$\pm$0.26    &49.78$\pm$0.43   &16.09$\pm$0.28    &57.38$\pm$0.46   &36.21$\pm$0.19   &32.64$\pm$0.71             \\
Mean Teacher    &70.09$\pm$1.60    &32.32$\pm$2.30  &9.19$\pm$0.19   &53.91$\pm$0.57  &35.83$\pm$0.24   &33.90$\pm$1.37             \\
UDA        &29.05$\pm$5.93   &8.82$\pm$1.08     &4.88$\pm$0.18   &33.13$\pm$0.22   &24.50$\pm$0.25    &6.64$\pm$0.17       \\
FixMatch    & 13.81$\pm$3.37    &5.07$\pm$0.65  &4.26$\pm$0.05   &28.29$\pm$0.11    &22.60$\pm$0.12      &6.25$\pm$0.33          \\
Dash        &8.93$\pm$3.11  &5.16$\pm$0.23     &4.36$\pm$0.11    &27.15$\pm$0.22        &21.88$\pm$0.07        &6.39$\pm$0.56             \\ \hline
MixMatch                 &47.54$\pm$11.50   &11.05$\pm$0.86  &6.42$\pm$0.10    &39.94$\pm$0.37     &28.31$\pm$0.33    &21.70$\pm$0.68        \\
+kNN-DV                    &47.18$\pm$8.31          &10.34$\pm$1.31            &5.91$\pm$0.37             &38.65$\pm$0.35               &27.73$\pm$0.28               &21.16$\pm$0.53             \\ 
\rowcolor{blue!10}+HDL           & 46.02$\pm$5.51         &9.70$\pm$0.72            &5.37$\pm$0.21              &36.91$\pm$0.24               &26.47$\pm$0.49               &20.58$\pm$0.61             \\ \hline
ReMixMatch    &19.10$\pm$9.64    &5.44$\pm$0.05    &4.72$\pm$0.13   &27.43$\pm$0.31       &23.03$\pm$0.56        &6.74$\pm$0.14        \\
+kNN-DV          &18.53$\pm$5.66           &4.87$\pm$0.24           &4.58$\pm$0.09             &26.85$\pm$0.46               &22.67$\pm$0.33               &6.49$\pm$0.21             \\ 
\rowcolor{blue!10}+HDL       &17.92$\pm$3.93           &4.32$\pm$0.17           &4.02$\pm$0.15             &26.13$\pm$0.57              &22.15$\pm$0.15               &6.02$\pm$0.08             \\ \hline
FreeMatch     &4.90$\pm$0.04      &4.88$\pm$0.18    &4.10$\pm$0.02     &26.47$\pm$0.20      &21.68$\pm$0.03   & 5.63$\pm$0.15            \\
+kNN-DV                    &\underline{4.86$\pm$0.13}           &4.29$\pm$0.43            &3.87$\pm$0.11            &\underline{25.96$\pm$0.79}        &21.34$\pm$0.52     &5.41$\pm$0.26             \\ 
\rowcolor{blue!10}+HDL    &\textbf{4.78$\pm$0.08}           &\textbf{3.91$\pm$0.35}            &\underline{3.65$\pm$0.07}             &\textbf{25.17$\pm$0.45}               &20.85$\pm$0.58 &\underline{5.13$\pm$0.18}             \\ \hline
DualMatch      &5.75$\pm$1.01    &4.89$\pm$0.52   &3.88$\pm$0.10    &27.08$\pm$0.23   &  20.78$\pm$0.15     &5.94$\pm$0.08             \\
+kNN-DV       &5.48$\pm$0.52           &4.23$\pm$0.29            &3.75$\pm$0.46              &26.59$\pm$0.62               &\underline{20.45$\pm$0.49}               &5.52$\pm$0.16             \\ 
\rowcolor{blue!10}+HDL        & 5.16$\pm$0.69         &\underline{3.96$\pm$0.41}            &\textbf{3.53$\pm$0.32}             &26.01$\pm$0.54               &\textbf{20.09$\pm$0.76}               &\textbf{5.07$\pm$0.23}             \\ \bottomrule \hline
\end{tabular}
\vskip -0.05in
\end{table*}

\textbf{Class unbalanced dataset: CIFAR-10-LT (IF = 50, 100, 200).} The degree of class imbalance is characterized by the imbalance factor (IF) \cite{paper25}. Assuming there are $C$ classes, with sample sizes of $n_1,n_2,\dots,n_C$, where $n_1\ge n_2\ge \cdots \ge n_C$, then IF is defined as $\frac{n_1}{n_C}$. CIFAR-10-LT \cite{paper24} is a long-tail version of CIFAR-10, and we select CIFAR-10-LT with IFs of $50$, $100$, and $200$ to verify HDL. Additionally, a labeling rate needs to be set. When the labeling rate is $10\%$, the most frequent class contains $500$ labeled samples and $4500$ unlabeled samples. In this work, the labeling rate is always set to $10\%$ \cite{paper10}.

\textbf{Implementation Details.} We used Wide ResNet-28-2 \cite{paper29}, Wide ResNet-28-2, Wide ResNet-28-8, and Wide ResNet-37-2 \cite{paper30} as the backbone networks on CIFAR-10, CIFAR-10-LT, CIFAR-100, and STL-10, respectively. For all semi-supervised learning methods, the training steps were set to $2^{20}$ on CIFAR-10 and stl-10, $2^{19}$ on CIFAR-100, and $2^{16}$ on CIFAR-10-LT \cite{paper10}. The remaining parameter settings are shown in Table \ref{table1}. We have selected MixMatch, ReMixMatch, FreeMatch, and DualMatch as our baselines and employed HDL to enhance their performance. Specifically, after the four semi-supervised models have completed training, we utilize HDL to re-label the unlabeled data and continue training for an additional $2^{5}$ steps.

\begin{table}[h]
\vskip -0.07in
\caption{Details of the experimental setup.}
\label{table1}
\vskip -0.08in
\centering  
\begin{small}
\renewcommand\arraystretch{1.13}
\setlength{\tabcolsep}{0.7pt} 
\begin{tabular}{l|cccc}
\hline \toprule
Dataset       & CIFAR-10 & CIFAR-100 & STL-10   & CIFAR-10-LT \\ \cline{2-5}
Model         & WRN-28-2 & WRN-28-8  & WRN-37-2 & WRN-28-2    \\ \cline{2-5}
Weight decay  & 0.0005   & 0.001     & 0.0005   & 0.0005      \\  \cline{2-5}
Batch size    & \multicolumn{4}{c}{64}                        \\ \cline{2-5}
Learning rate & \multicolumn{4}{c}{0.03}                      \\ \cline{2-5}
SGD momentum  & \multicolumn{4}{c}{0.9}                       \\ \cline{2-5}
EMA decay     & \multicolumn{4}{c}{0.999}                     \\  \cline{2-5}
$k$     & 3 &3 &3  &3                     \\  
\bottomrule  \hline
\end{tabular}
\end{small}
\vskip -0.05in
\end{table}

\textbf{Compared Methods.} Similar to HDL, we enhanced MixMatch \cite{paper3}, ReMixMatch \cite{paper4}, FreeMatch \cite{paper9}, and DualMatch \cite{paper10} using kNN-DV (Section \ref{sec4.2}) and compared them with HDL. Additionally, we contrasted our method with common semi-supervised models, including $\Pi$-Model \cite{paper26}, Pseudo-Labeling \cite{paper14}, Mean Teacher \cite{paper27}, UDA \cite{paper28}, FixMatch \cite{paper6}, and Dash \cite{paper7}.

\subsection{Setting of the hyperparameter $k$}
\label{sec5.3}
The trained semi-supervised model to be improved is used to extract the image embeddings, and then the $k$ is chosen based on the methodology presented in Section \ref{sec4.4}. We present the $k$ values on each dataset in the last row of Table \ref{table1}.

\subsection{Results on Class-Balanced Datasets}
\label{sec5.4}

Whether it is a consistency regularization model or a pseudo-labeling model, their common goal is to optimize the model, making its representations of similar samples more consistent and similar, thereby enhancing performance. Observing the results of kNN-DV and HDL in Table \ref{table2}, it is evident that they significantly improve the performance of existing semi-supervised learning methods across various datasets. Experimental results indicate that image embeddings generated by adequately trained semi-supervised learning models are more reliable than confidence predictions. Therefore, we emphasize the importance of focusing on pseudo-label semi-supervised models based on image embeddings.

Specifically, on the CIFAR-10 with $40$ labels, HDL improves the performance of MixMatch and ReMixMatch by $\textbf{1.52\%}$ and $\textbf{1.18\%}$, respectively. On the CIFAR-10 with $250$ labels, HDL yields performance gains of $\textbf{1.35\%}$, $\textbf{1.12\%}$, $\textbf{0.97\%}$, and $\textbf{0.93\%}$ for MixMatch, ReMixMatch, FreeMatch, and DualMatch, respectively, with the enhanced FreeMatch achieving \textbf{state-of-the-art performance}. Furthermore, on the CIFAR-10 dataset with $4000$ labels, HDL enhances the performance of MixMatch by $\textbf{1.05\%}$, and DualMatch + HDL achieves \textbf{optimal performance}.
For the CIFAR-100 and STL-10, HDL consistently enhances the performance of existing methods comprehensively. In contrast, the performance of k-NN-DV is relatively mediocre. On the CIFAR-10 dataset with $4000$ labels and the CIFAR-100 dataset with $10000$ labels, k-NN-DV only brings about an average performance improvement of approximately $0.5\%$ compared to existing methods.
\textbf{It is important to note} that our approach does not require any changes to the existing method, only relabeling and continued training of the original network.

\subsection{Results on Class-Imbalanced Datasets}
\label{sec5.5}

The evaluation results on CIFAR-10-LT, which features multiple imbalance factors, are shown in Table \ref{table3}. Compared to scenarios with class balance, HDL exhibits more pronounced performance on long-tailed datasets. This aligns with our expectations, as image embedding-based methods in long-tailed scenarios can reduce model bias. Specifically, at an Imbalance Factor (IF) of $50$, HDL achieved performance gains of $\textbf{1.7\%}$, $\textbf{1.5\%}$, and $\textbf{1.4\%}$ for MixMatch, FixMatch, and DualMatch, respectively. With an IF of $100$, HDL improved the performance of MixMatch by $\textbf{1.5\%}$. However, the benefits provided by HDL diminished when the IF was increased to $200$, likely due to a decrease in the number of tail-class samples participating in voting. Moreover, HDL's performance comprehensively surpassed that of kNN-DV, further validating the soundness of our approach.

\begin{table}[h]
\vskip -0.05in
\caption{Error rates (mean $\pm$ std $\%$) on CIFAR-10-LT. The best results are highlighted in \textbf{bold}, and the second-best results are \underline{underlined}. Note that we evaluated our approach using $5$ experiments with different random seeds.}
\label{table3}
\vskip -0.1in
\centering  
\renewcommand\arraystretch{1.03}
\setlength{\tabcolsep}{5.7pt} 
\begin{tabular}{l|ccc}
\hline \toprule
Method          & IF=50 & IF=100 & IF=200 \\ \hline
Pseudo-Labeling & 47.5$\pm$0.74  & 53.5$\pm$1.29   & 58.0$\pm$1.39   \\
Mean Teacher    & 42.9$\pm$3.00  & 51.9$\pm$0.71   & 54.9$\pm$1.28   \\ \hline
MixMatch        & 30.9$\pm$1.18  & 39.6$\pm$2.24   & 45.5$\pm$1.87   \\
+kNN-DV                 &29.8$\pm$1.27       &38.8$\pm$2.05        &44.8$\pm$1.72        \\
\rowcolor{pink!40}+HDL                  &29.2$\pm$0.84       &38.1$\pm$1.69        &44.3$\pm$1.25        \\ \hline
FixMatch        & 20.6$\pm$0.65  & 33.7$\pm$1.74   & 40.3$\pm$0.74   \\
+kNN-DV                  &19.7$\pm$0.92       &32.9$\pm$1.31        &39.9$\pm$1.19        \\
\rowcolor{pink!40}+HDL                  &19.1$\pm$1.13       &32.1$\pm$1.52        &39.6$\pm$0.82        \\ \hline
DualMatch       & 19.0$\pm$0.82  & 28.3$\pm$1.38   & 37.3$\pm$0.39   \\
+kNN-DV                  &\underline{18.3$\pm$1.08}       &\underline{27.5$\pm$0.96}        &\underline{36.8$\pm$0.68}        \\
\rowcolor{pink!40} +HDL                  &\textbf{17.6$\pm$1.36}       &\textbf{27.0$\pm$1.02}        &\textbf{36.7$\pm$0.45}        \\  \bottomrule  \hline
\end{tabular}
\vskip -0.1in
\end{table}

\subsection{Additional analysis of kNN-DV and HDL}
\label{sec5.6}

\begin{figure}[t]
\centering
\includegraphics[width=3.4in]{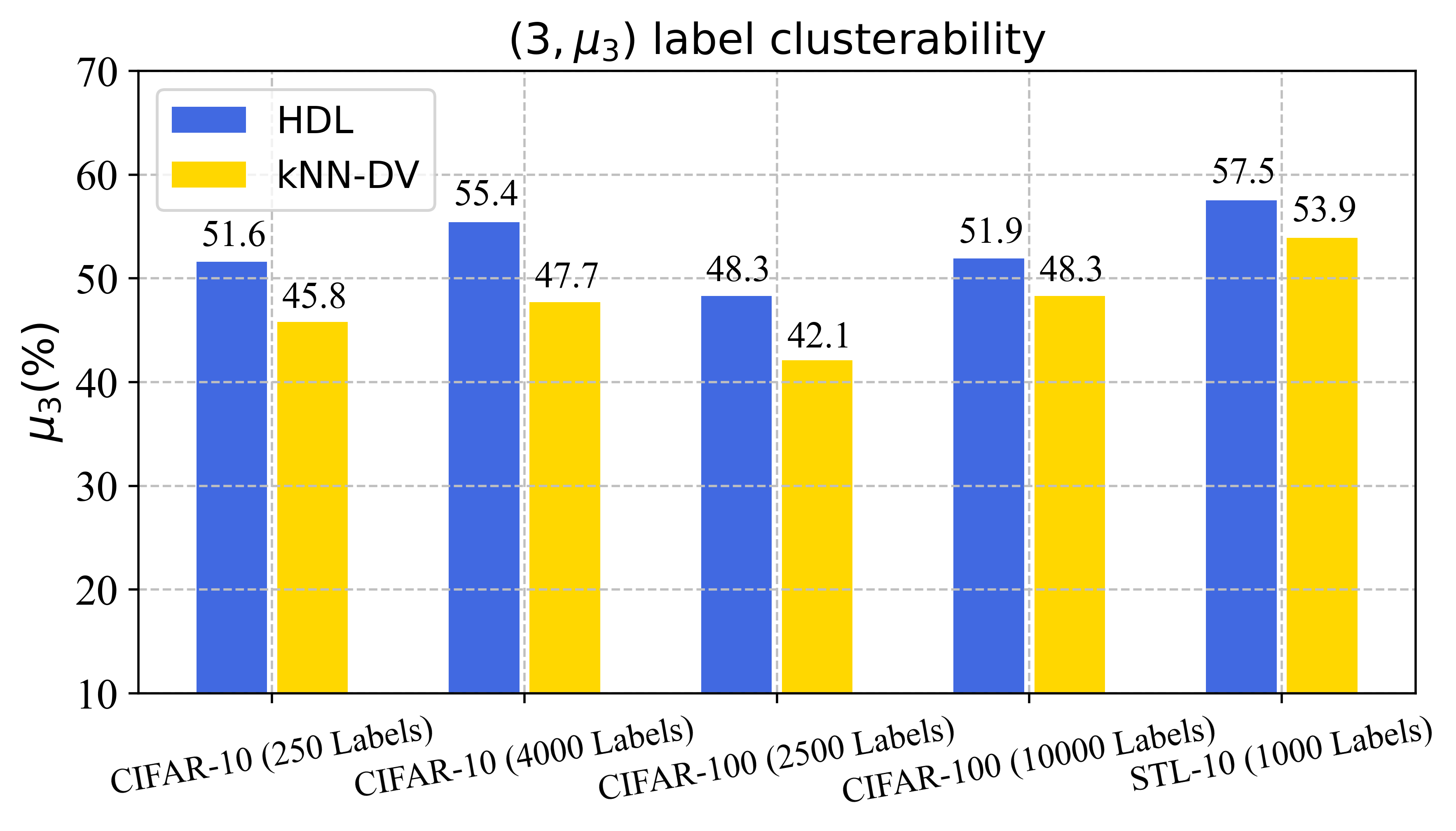}
\vskip -0.05in
\caption{The probability $\mu_3$ that clustrability holds on CIFAR-10, CIFAR-100 and STL-10 for $k = 3$.}
\label{fig3}
\vskip -0.02in
\end{figure}


Our analysis in Section \ref{sec4.2} suggests that kNN-DV, by not accounting for unlabeled samples, potentially raises the requirements for the clusterability of the image embedding sets to be valid. In this section, we set the hyperparameter $k$ for both kNN-DV and HDL to $3$ to quantify and analyze the aforementioned shortcomings of kNN-DV.
In our investigation of the CIFAR-10, CIFAR-100, and STL-10 datasets, we found that the probability of clusterability for kNN-DV is consistently lower than that for HDL across all three datasets. To enhance clusterability, it becomes necessary to reduce the number of nearest neighbors. However, reducing the number of nearest neighbors can compromise the reliability of the algorithm. Therefore, it can be concluded that HDL theoretically surpasses kNN-DV in performance, a finding that is also corroborated by our experimental results.

\subsection{HDL for data processing modules}
\label{sec5.7}

\begin{figure}[h]
\vskip -0.05in
\centering
\includegraphics[width=3.34in]{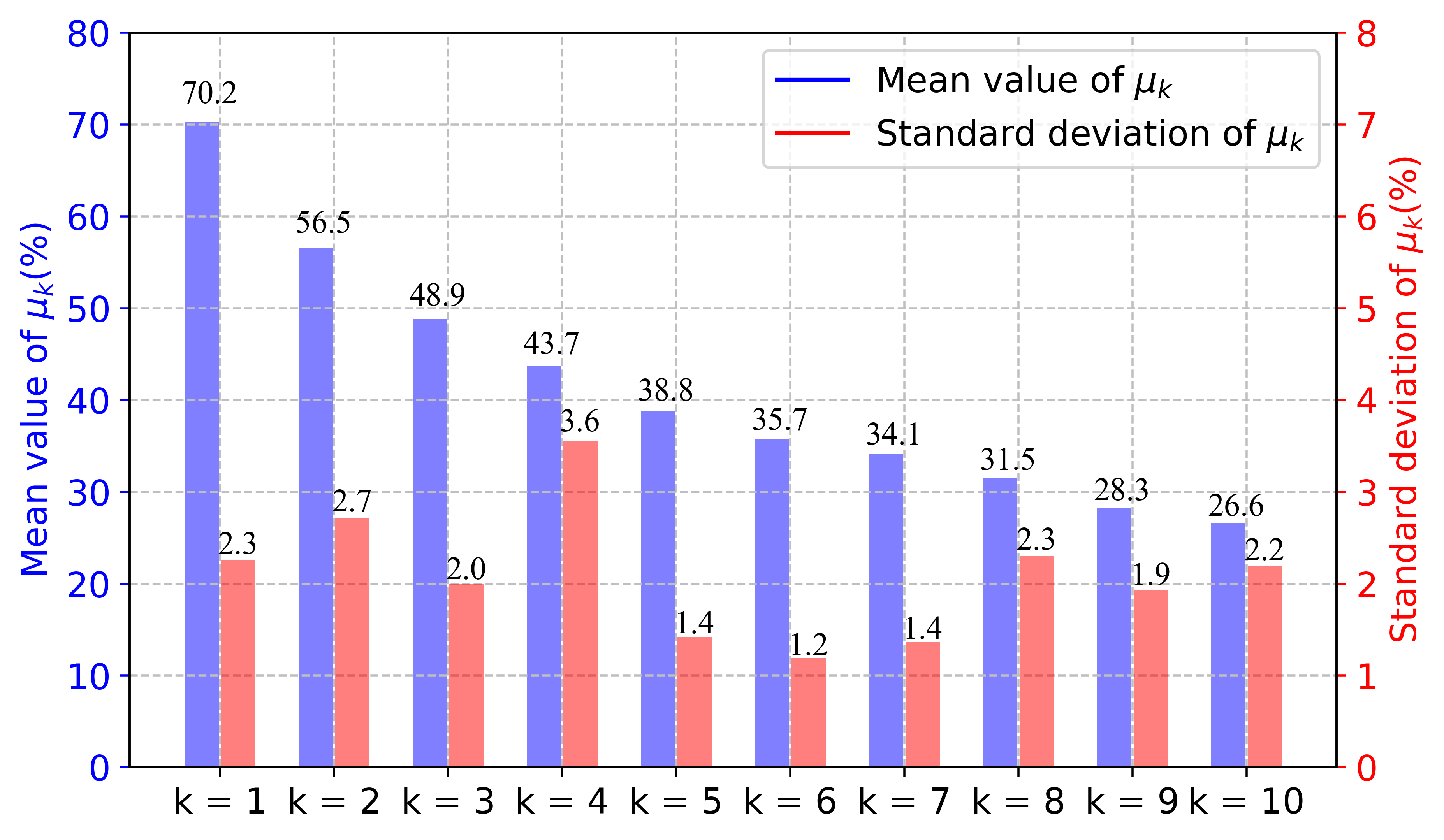}
\vskip -0.05in
\caption{Under different values of $k$, the means and standard deviations of the probabilities that the embedding sets for the four image datasets satisfy clusterability. Please refer to the \textcolor{blue}{blue vertical axis} for the magnitude of the mean and the \textcolor{red}{red vertical axis} for the magnitude of the standard deviation.}
\label{fig2}
\vskip -0.15in
\end{figure}

When HDL is employed as a general data processing module, CLIP can be used to extract image embeddings. Given that CLIP is a universal image encoder, we hypothesize that the probability of the extracted embedding sets satisfying clusterability is roughly the same across different image datasets. We assessed the probability of label clusterability, $\mu_k$, under different values of $k$ on the mini-ImageNet, Clothing-1M, CIFAR-100, and Caltech 101 datasets. Subsequently, we calculated and plotted the mean and variance of $\mu_k$ in Figure \ref{fig2}. The experimental results reveal that the variance of $\mu_k$ is very small across all four datasets. This validates our viewpoint and implies that in practice, it is not necessary to statistically evaluate $\mu_k$ each time; instead, a reliable empirical value can be used as a substitute.

\section{Conclusion}
\label{sec6}
This work originates from the perspective that embeddings are more reliable than confidence in predictions. It introduces the Embedding-based Hierarchical Dynamic Labeling algorithm and significantly enhances existing semi-supervised models. This advancement holds the potential to foster improvements in the pseudo-label generation paradigm.

\bibliographystyle{named}
\bibliography{ijcai23}

\end{document}